\pdfoutput=1

\documentclass[11pt]{article}

\usepackage{fontawesome}
\usepackage[final]{EMNLP2023}

\usepackage{times}
\usepackage{latexsym}
\usepackage{spverbatim}

\usepackage[T1]{fontenc}

\usepackage[utf8]{inputenc}

\usepackage{microtype}

\usepackage{inconsolata}
\usepackage{microtype}
\usepackage{graphicx}
\usepackage{multirow} 
\usepackage{booktabs} 
\usepackage{longtable}
\usepackage{kotex}
\usepackage[dvipsnames]{xcolor}
\usepackage{changepage}
\usepackage{makecell}

\usepackage[table]{xcolor}
\definecolor{ltblue}{RGB}{220,230,241}
\definecolor{ltred}{RGB}{241,220,220}
\title{Evaluating the Pre-Consultation Ability of LLMs\\ using Diagnostic Guidelines}

\author{Jean Seo$^{1}$, Gibaeg Kim$^{1}$, Kihun Shin$^{3}$, Seungseop Lim$^{1}$, \\ \textbf{Hyunkyung Lee$^{1}$, Wooseok Han$^{1}$, Jongwon Lee$^{4}$, Eunho Yang$^{1,2}$}  \\
    $^{1}$AITRICS \qquad
    $^{2}$KAIST \\
    $^{3}$Severance Hospital, Yonsei University \\
    $^{4}$College of Medicine, The Catholic University of Korea \\
         \texttt{\{jeanseo\}@aitrics.com}}

\begin{document}
\maketitle

\begin{abstract}
We introduce \textbf{EPAG}, a benchmark dataset and framework designed for \textbf{E}valuating the \textbf{P}re-consultation \textbf{A}bility of LLMs using diagnostic \textbf{G}uidelines. LLMs are evaluated directly through HPI-diagnostic guideline comparison and indirectly through disease diagnosis. In our experiments, we observe that small open-source models fine-tuned with a well-curated, task-specific dataset can outperform frontier LLMs in pre-consultation. Additionally, we find that increased amount of HPI (History of Present Illness) does not necessarily lead to improved diagnostic performance. Further experiments reveal that the language of pre-consultation influences the characteristics of the dialogue. By open-sourcing our dataset and evaluation pipeline on \url{https://github.com/seemdog/EPAG}, we aim to contribute to the evaluation and further development of LLM applications in real-world clinical settings.

\end{abstract}

\section{Introduction}


Large Language Models (LLMs) are increasingly integrated into clinical applications, transforming healthcare industry by automating various tasks \citep{clinical_chatbot, zhou2024survey, thirunavukarasu2023large, wang2025surveyllmbasedagentsmedicine}. One  example is pre-consultation, where LLMs assist history-taking \citep{10803352, article} and decision-making \citep{samieegeneral, li2024beyond}. However, it is crucial to acknowledge the significant risks involved. Erroneous outputs can result in severe adverse consequences such as mistreatment or incorrect drug prescription, highlighting the necessity of rigorous evaluations \citep{kim2025medicalhallucinationsfoundationmodels, ullah2024challenges}.

\begin{figure*}[t]
    \centering
    \includegraphics[width=0.95\textwidth]{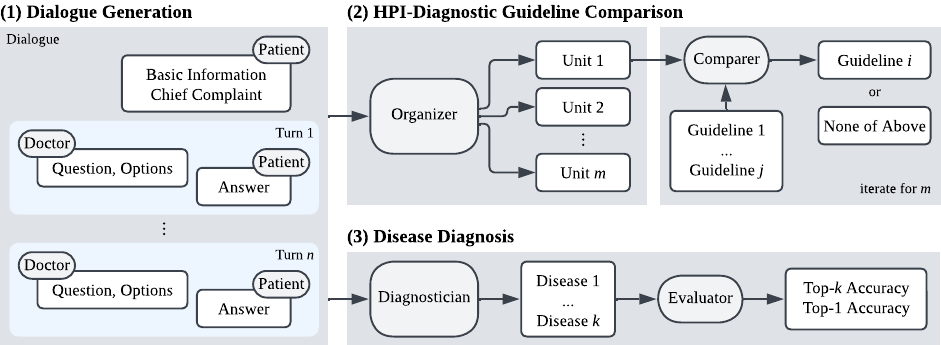}
    \caption{EPAG pipeline. \textbf{(1) Dialogue Generation}: The patient-agent acts as a patient given a specific profile, while the doctor-agent conducts a pre-consultation using only the basic information and chief complaint. After \textit{n} turns, the doctor-agent is assessed through two tasks: \textbf{(2) HPI-Diagnostic Guideline Comparison}, where the organizer model extracts HPI units and the comparer model determines which of the diagnostic guidelines is most relevant, and \textbf{(3) Disease Diagnosis}, where the dialogue is given to a separate diagnostician-agent for diagnosis.
    }
    \label{fig:pipeline}
\end{figure*}

We propose \textbf{EPAG} (\textbf{E}valuating the \textbf{P}re-consultation \textbf{A}bility of LLMs using diagnostic \textbf{G}uidelines), a benchmark dataset and evaluation pipeline specifically designed for pre-consultation. Given basic patient information, such as age, sex, and chief complaints, pre-consultation models ask questions to elicit symptoms related to potential diagnoses. EPAG benchmark dataset comprises 520 patient profiles, spanning 26 diseases, 10 ICD-11 chapters, 10 primary specialties, and 22 secondary specialties, along with pre-defined diagnostic guidelines. In EPAG, the pre-consultation dialogue is evaluated through two tasks: (1) HPI-Diagnostic Guideline Comparison, and (2) Disease Diagnosis. In our experiments, eleven LLMs are evaluated across various numbers of dialogue turns.

The main contributions of our work are:

\begin{itemize} 
\item Developing a systematic framework and constructing a high-quality dataset for evaluating the clinical pre-consultation ability of LLMs. 
\item Open-sourcing the dataset and pipeline. 
\item  Implementing targeted experiments and sharing the results with in-depth analysis.
\end{itemize}

\section{Related Work}

\subsection{Medical LLMs in Clinical Applications}
Existing clinical chatbot applications include HuatuoGPT \citep{zhang-etal-2023-huatuogpt}, ChatDoctor \citep{li2023chatdoctormedicalchatmodel}, MedChatZH \citep{tan2024medchatzh}, MedAide \citep{basit2024medaideleveraginglargelanguage}, and MILD Bot \citep{kim-etal-2024-mild}. Other medical LLM applications not limited to chatbots are \citet{kumichev2024medsyn, zhang2024llmbasedmedicalassistantpersonalization, wiest2024anonymizing, ghosh2024clipsyntel, waisberg2024large}. LLMs have demonstrated diagnostic accuracy comparable to that of physicians in certain contexts \citep{qian2021pre}, with existing works primarily focusing on final diagnostic outcomes \citep{mcduff2023accuratedifferentialdiagnosislarge, singhal2023expertlevelmedicalquestionanswering, tu2024conversationaldiagnosticai}. However, research on patient information collection during LLM pre-consultation remains limited. To address this, we propose a fine-grained framework that evaluates LLM pre-consultation capabilities.

\subsection{Evaluation of Medical LLMs}

Multiple-choice QA is widely used for medical evaluation, as demonstrated by Med-HALT \citep{pal2023medhaltmedicaldomainhallucination}, MedMCQA \citep{pal2022medmcqalargescalemultisubject}, PubMedQA \citep{jin-etal-2019-pubmedqa}, and KoreMedMCQA \citep{kweon2024kormedmcqamultichoicequestionanswering}. However, it is insufficient for assessing real-world clinical conversational abilities \citep{bedi2024systematic, chen2024evaluating}. More sophisticated evaluation frameworks in the clinical domain have been proposed, including MEDIC \citep{kanithi2024mediccomprehensiveframeworkevaluating}, LLM-Mini-CEX \citep{shi2023llmminicexautomaticevaluationlarge}, CRAFT-MD \citep{johri2025craftmd}. Other evaluation benchmarks regarding disease diagnosis include works by \citet{hou2024msdiagnosisbenchmarkevaluatinglarge}, \citet{zhu2025diagnosisarenabenchmarkingdiagnosticreasoning}, \citet{bhasuran2025preliminary}, \citet{delaunay2024evaluating},  \citet{sarvari2025rapidly}, \citet{reese2025systematic}, \citet{gaber2025evaluating}. While \citet{winstonmedical} and \citet{fast-2024-amega-llm} propose evaluation pipelines for pre-consultation, their dataset coverage is limited and peripheral. 

\begin{figure*}[t]
    \centering
    \includegraphics[width=0.96\textwidth]{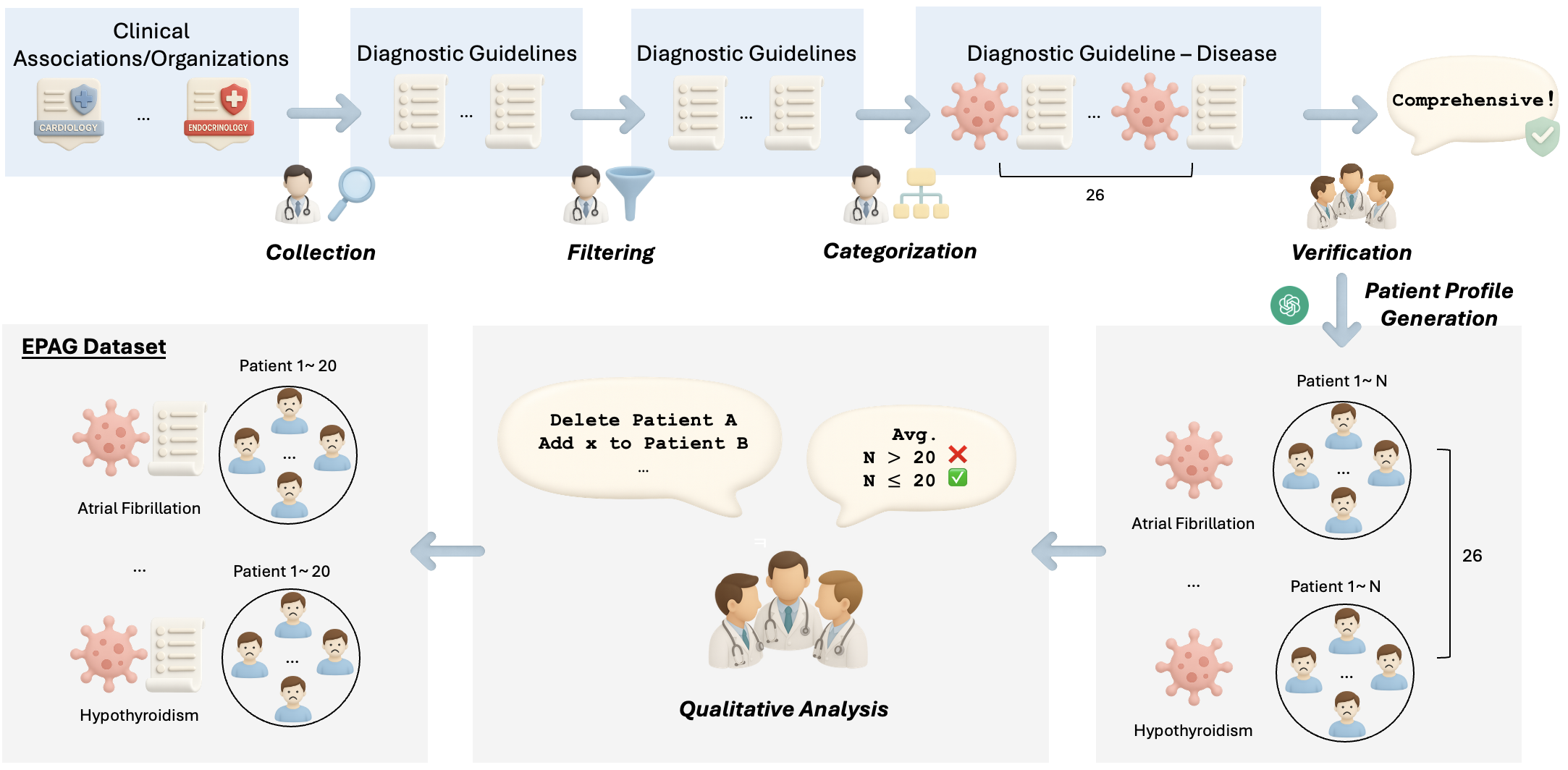}
    \caption{EPAG benchmark dataset construction process. Expert clinicians collect all possible diagnostic guidelines of diseases from credible clinical sources. They then filter diseases based on whether they can be reasonably diagnosed through consultation alone and sufficiently common to ensure unbiased evaluation. Next, clinicians verify that the disease list is comprehensive enough to serve as a generalizable evaluation set. Using the finalized list, synthetic patient profiles are generated and finalized through qualitative analysis by clinicians.}
    \label{fig:dataset}
\end{figure*}

\section{EPAG Benchmark}

We assess pre-consultation models designed to collect as much relevant information as possible from the patient, including symptoms, family history, and other factors, referred to as the History of Present Illness (HPI). This section covers the tasks, dataset construction process, and evaluation pipeline of EPAG.

\subsection{Evaluation Tasks}
\label{sec:task}

As Figure \ref{fig:pipeline} demonstrates, we propose a two-tiered evaluation framework based on the collected HPI.

\subsubsection{HPI-Diagnostic Guideline Comparison}

For direct evaluation, we focus on how effectively the models capture information necessary for accurate disease identification. The evaluation process involves pre-consultation simulation with a patient-agent exhibiting symptoms of a specific disease and a doctor-agent, which is the subject of evaluation. During this interaction, the doctor-agent asks questions and provides multiple options for the patient-agent to choose from. The HPI collected is then compared against a set of diagnostic guidelines for the specific disease. The diagnostic guidelines represent a collection of essential information for diagnosing a particular disease, curated by human clinicians from trusted sources with further details in Section \ref{sec:diagnostic_guideline}.

\subsubsection{Disease Diagnosis}

For indirect evaluation, we assess how well the collected HPI supports accurate diagnoses when provided to a separate diagnostic model. While this is not a direct evaluation of the HPI extracted by LLMs, it is a crucial assessment as one of the eventual goals of LLM pre-consultation is to assist in correct diagnosis and treatment.

\subsection{Dataset}
Figure \ref{fig:dataset} shows the dataset construction process.


\subsubsection{Diagnostic Guideline}
\label{sec:diagnostic_guideline}

To evaluate whether each dialogue turn elicits meaningful patient information for diagnosis, we construct a gold-label diagnostic guideline dataset. The following steps are implemented by professional clinicians based on credible clinical associations and organizations in Appendix \ref{appendix:source}:  
(1) collect diagnostic guidelines with explicit references;  
(2-1) filter diseases that are diagnosable through consultation alone, without reliance on physical exams, X-ray or MRI;  
(2-2) exclude diseases that are too rare.  
As exemplified in Appendix \ref{appendix:guideline}, each diagnostic guideline specifies key symptoms, ancillary symptoms, family history, and other relevant risk factors. Each feature is assigned a weight of either \textit{high} or \textit{medium}.

\subsubsection{Disease}

As our primary goal is to evaluate language models rather than multi-modal models, we focus on diseases that can be differentiated without reliance on other examination results. Through extensive discussions with clinicians, we identify 26 such diseases spanning 10 primary specialties and 22 secondary specialties. To ensure that the selection of 26 diseases provides sufficient clinical generalizability, clinicians classify them according to the International Classification of Diseases, 11th Revision (ICD-11) \footnote{https://icd.who.int/en}. This categorization confirms that the included diseases span a broad range of conditions across 10 ICD-11 chapters, as shown in Table \ref{tab:disease_all}, indicating that the dataset covers a clinically diverse and representative scope of diseases that can be reasonably differentiated through history-taking. Each disease is systematically assigned to both primary and secondary specialties following established clinical criteria in Appendix \ref{appendix:disease}, reflecting the multidisciplinary nature of real-world patient care.

\subsubsection{Patient Profile}

We generate diverse patient profiles using OpenAI o3-mini \footnote{https://openai.com/}. Expert clinicians then conduct a qualitative review to ensure (i) sufficient diversity across profiles and (ii) adequate clinical detail to support realistic patient–doctor interactions. To minimize bias in the synthetic dataset, we retain 20 profiles per disease, yielding a total of 520 profiles. Each profile contains demographic and clinical information such as age, sex, height, weight, and relevant medical history, representing realistic patient cases. Each patient profile is used to assign a role to the patient-agent, which then interacts with the doctor-agent, simulating realistic scenarios. A sample profile and diversity of patient group can be found in Table \ref{tab:sample_patient} and Figure \ref{fig:diversity} respectively.

\begin{table}[ht]
\centering
\resizebox{\columnwidth}{!}{%
\begin{tabular}{ccccc}
\toprule[1.3pt]
\multirow{2}{*}{\textbf{Model}}
 & \multicolumn{2}{c}{\textbf{\makecell{HPI‐Diagnostic Guideline \\ Comparison Score}}}
 & \multicolumn{2}{c}{\textbf{\makecell{Disease Diagnosis \\ Accuracy}}}\\
\cmidrule(lr){2-3} \cmidrule(lr){4-5}
 & \textbf{Not Weighted}
 & \textbf{Weighted}
 & \textbf{Top‐1}
 & \textbf{Top‐\textit{k}} \\
\midrule[1.3pt]

\specialrule{0pt}{1.1pt}{1.1pt}

\textbf{Human Expert} & 4.35 & 7.29 & 68.24 & 80.65 \\
\midrule

\specialrule{0pt}{1.1pt}{1.1pt}
\multicolumn{5}{c}{\textbf{LLMs}}\\
\specialrule{0pt}{1.1pt}{1.1pt}
\midrule

\textbf{GPT‐4.1}
 & \cellcolor{ltblue}\textbf{4.82}
 & \cellcolor{ltblue}\textbf{8.12}
 & \cellcolor{ltblue}\textbf{74.56}
& \cellcolor{ltblue}\textbf{83.81} \\

\textbf{GPT‐4.1‐mini}
 & \cellcolor{ltblue}4.46
 & \cellcolor{ltblue}7.64
 & \cellcolor{ltblue}69.15
 & \cellcolor{ltblue}81.36\\

\textbf{GPT‐4o}
 & \cellcolor{ltblue}4.39
 & \cellcolor{ltblue}7.59
 & \cellcolor{ltblue}69.23
 & \cellcolor{ltblue}81.35\\

\textbf{GPT‐4o‐mini}
 & \cellcolor{ltblue}4.46
 & \cellcolor{ltblue}7.75
 & \cellcolor{ltred}64.62
 & \cellcolor{ltred}79.62\\

\textbf{Claude‐3.7‐Sonnet}
 & \cellcolor{ltblue}4.59
 & \cellcolor{ltblue}\textbf{8.12}
 & \cellcolor{ltblue}69.23
 & \cellcolor{ltblue}82.31\\

\textbf{Claude‐3.5‐Sonnet}
 & \cellcolor{ltblue}4.62
 & \cellcolor{ltblue}8.05
 & \cellcolor{ltblue}72.69
 & \cellcolor{ltblue}81.35\\

\textbf{Claude‐3.5‐Haiku}
 & \cellcolor{ltblue}4.58
 & \cellcolor{ltblue}7.84
 & \cellcolor{ltred}65.38
 & \cellcolor{ltblue}80.77\\

\textbf{Phi-3.5-mini}
 & \cellcolor{ltred}3.91
 & \cellcolor{ltred}6.88
 & \cellcolor{ltred}61.82
 & \cellcolor{ltred}78.84\\

 \textbf{Llama‐3.2‐3B}
 & \cellcolor{ltred}3.87
 & \cellcolor{ltred}6.8
 & \cellcolor{ltred}58.14
 & \cellcolor{ltred}72.09\\

\textbf{Qwen2.5‐7B}
 & \cellcolor{ltred}3.74
 & \cellcolor{ltred}6.51
 & \cellcolor{ltred}58.46
 & \cellcolor{ltred}76.54\\

\textbf{Medgemma-4B} \faStethoscope
 & \cellcolor{ltred}4.19
 & \cellcolor{ltred}7.22
 & \cellcolor{ltred}65.93
 & \cellcolor{ltblue}82.31\\

\bottomrule[1.3pt]
\end{tabular}%
}
\caption{HPI–diagnostic guideline comparison scores and disease diagnosis accuracies for eleven models, alongside a human expert baseline, over five-turn dialogues. Results exceeding the human baseline are shaded in blue, and those below in red. Stethoscope (\faStethoscope) denotes the medically fine-tuned model.}
\label{tab:result}
\end{table}

\subsection{Evaluation Framework}
\label{sec:framework}

Supposing pre-consultation models that ask questions and provide options to choose from, [Question, Options, Answer] triplets are utilized throughout evaluation.

\subsubsection{HPI-Diagnostic Guideline Comparison Score}
\label{sec:hpi_gl}

\noindent\textbf{(1) Response Generation}\\
The doctor-agent is provided with the chief complaint and basic information, including age, sex, height, weight, then generates questions and options. The patient-agent is provided with the full patient profile, and asked to select the appropriate 
 option with the prompt in Table \ref{tab:patient_prompt}. This process is iterated for \textit{n} times. 
 
\noindent\textbf{(2) Organization}\\
After \textit{n} turns of pre-consultation, the [Question, Options, Answer] triplets are organized into individual units, each representing a single piece of clinical information, by an organizer model, using the prompt in Table \ref{tab:organizer_prompt}. This step is crucial because, in the next phase, we compare each unit against pre-defined diagnostic guidelines to assess whether it matches any. Since a single [Question, Options, Answer] triplet may contain multiple pieces of information, separating them into individual units ensures more accurate comparison. 
For example: 

\begin{adjustwidth}{-2mm}{0cm}
\begin{quote}
\textbf{Question:} 
Are there any other symptoms that occur with chest tightness? \\
\textbf{Options:} 
Shortness of breath or difficulty breathing,
A feeling of a racing heart,
Cold sweats,
Dizziness,
Vomiting or nausea\\
\textbf{Answer:} 
Shortness of breath or difficulty breathing
\end{quote}
\end{adjustwidth}

The number of organized units should be five, not one: (1) Patient has shortness of breath or difficulty breathing, (2) Patient does not have a racing heart, (3) nor cold sweats, (4) nor dizziness, (5) nor vomiting or nausea. In differential diagnosis, the absence of symptoms is as significant as their presence, so the unselected options are treated as separate units. Additionally, to avoid duplicating scores for redundant questions, we deduplicate the information extracted during the organization step. An example is provided in Appendix \ref{appendix:unit}.

\noindent\textbf{(3) Comparison}\\
Next, we use a comparer model with the prompt in Table \ref{tab:comparer_prompt} to match each unit with the most relevant diagnostic guideline. If a unit does not match any of the guidelines, the comparer model is instructed to respond with "None of Above." As illustrated in Figure \ref{fig:pipeline}, for each of the \textit{m} units, the comparer performs the comparison process.

\noindent\textbf{(4) Score Calculation}\\
The final score for the pre-consultation dialogue is calculated by awarding 1 point if the unit corresponds to a guideline and 0 point for "None of Above." Since some diagnostic guidelines may be more influential in diagnosing or ruling out certain diseases than others, we also compute a weighted score. Human expert clinicians assign each guideline a significance level of medium or high, as shown in Appendix \ref{appendix:guideline}. A unit corresponding to a medium-significance guideline earns 1 point, while a high-significance guideline earns 2 points. Both versions of the score are calculated for each patient and averaged across 520 datasets to determine the final score for each doctor-agent.

To verify the reliability of our evaluation pipeline, we conduct a human comparison. For each disease, one is randomly sampled for each disease and evaluated by a human clinician using the same pipeline. After performing an F-test (p > 0.05) to ensure equal variances, a T-test confirms that the two sets of scores are statistically similar (p > 0.05).

\subsubsection{Disease Diagnosis Accuracy}

For indirect evaluation of the pre-consultation dialogue, we use an independent diagnostician-agent with the prompt in Table \ref{tab:diagnostician_prompt}. To account for multiple names for the same disease, we consider the prediction correct if the model identifies a parent or child concept of the gold label disease. We employ an evaluator model using the prompt in Table \ref{tab:evaluator_prompt} to determine if the predicted disease matches the gold label. 

\section{Experiments}

We evaluate eleven models as the doctor-agent, including four from OpenAI \footnote{https://openai.com/}, three from Anthropic \footnote{https://www.anthropic.com/}, and four open-source LLMs, one of which is medically fine-tuned, and compare their performance to a human baseline. For the human baseline, human clinicians go through the same pre-consultation process as the doctor-agents, while the rest of the pipeline remains unchanged. Figure \ref{fig:human_screen} shows the user interface used by human clinicians. The resulting pre-consultation dialogues are then evaluated using our proposed pipeline. In this experiment, all other components in the pipeline use GPT-4o-mini, with distinct prompts assigned to each role (patient, organizer, comparer, diagnostician, evaluator). To ensure reproducibility, we fix the random seed and set the temperature of each agent to 0. The only variable is the doctor-agent model.


\begin{figure}[t]
    \centering
    \includegraphics[width=0.95\columnwidth]{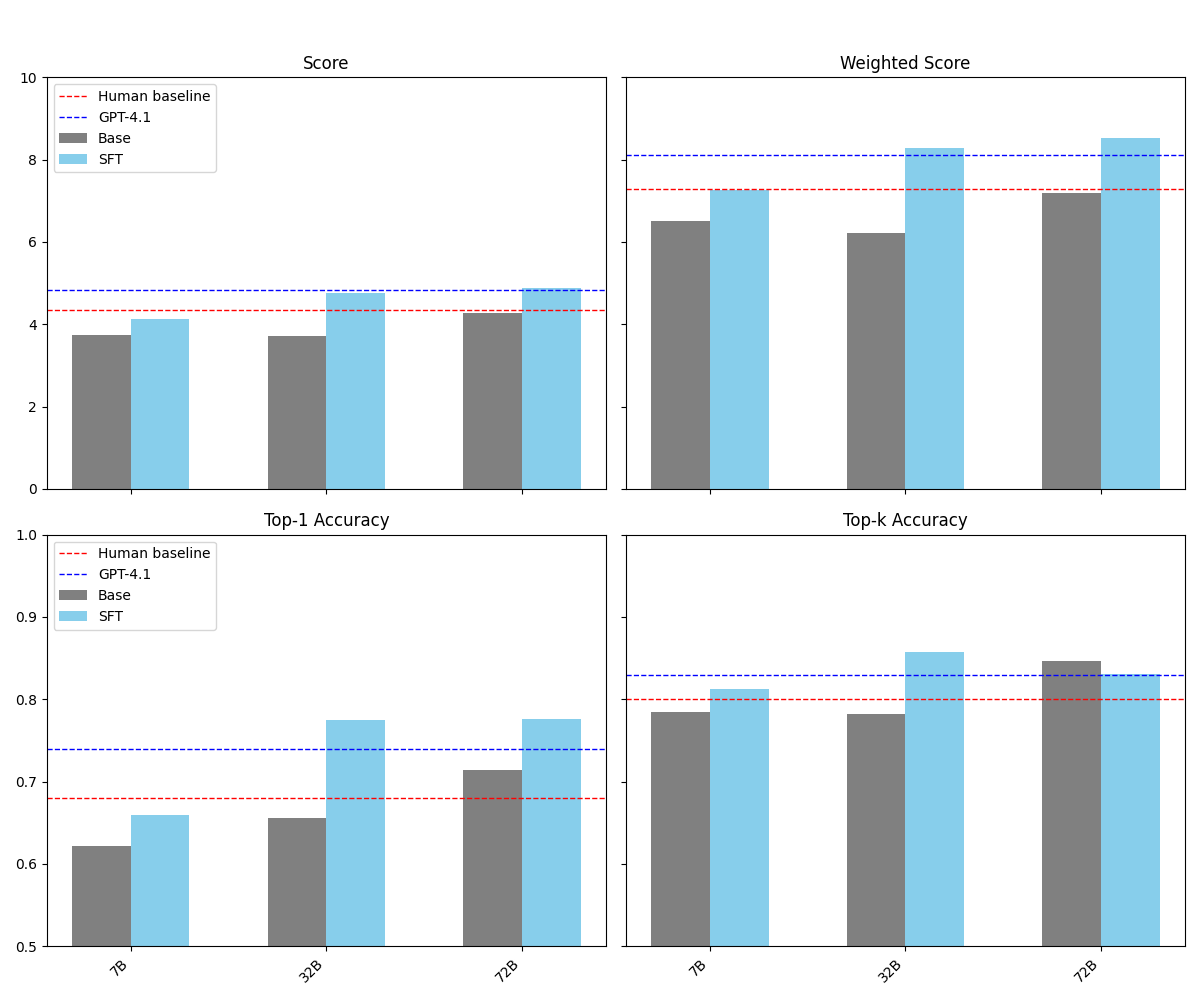}
    \caption{Performance of Qwen-2.5 models (7B, 32B, 72B) before (grey) and after (blue) SFT. Red horizontal line marks human clinician performance, and blue marks GPT-4.1 performance—the strongest model.}
    \label{fig:sft}
\end{figure}

\begin{figure*}[t]
    \centering
    \includegraphics[width=0.95\textwidth]{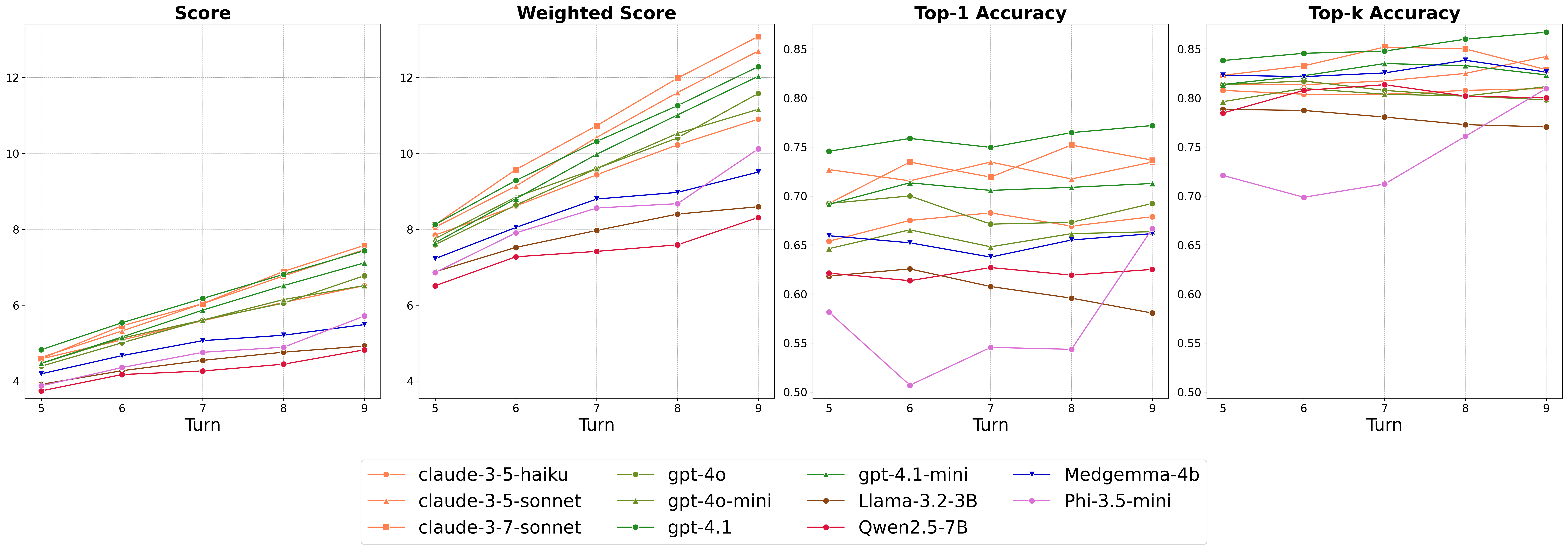}
    \caption{EPAG results across eleven models with number of dialogue turns ranging from five to nine. }
    \label{fig:epag}
\end{figure*}

\section{Result and Analysis}

As shown in Table \ref{tab:result}, in five turn dialogues, GPT-4.1 attains the highest performance across all metrics, tying with Claude-3.7-Sonnet on the weighted HPI-diagnostic guideline comparison score. Qwen2.5-7B and Llama-3.2-3B perform worst overall. The human baseline places above all open-source models, but below every proprietary LLM. Contrary to our intuition that medical fine-tuning would elicit decent performance, Medgemma-4B underperforms the human baseline. A plausible explanation is that Medgemma-4B is fine-tuned primarily on existing medical tasks, which may have weakened its instruction-following ability on unseen tasks like pre-consultation. 
We conduct a series of additional experiments, providing several important takeaways.

\noindent\textbf{Model size does not guarantee performance.} \\
Larger or more expensive models are expected to outperform  their smaller counterparts across most tasks. This holds true in the GPT-4.1 family, where GPT-4.1 exceeds GPT-4.1-mini on all four metrics. However, GPT-4o-mini outperforms GPT-4o on HPI-diagnostic guideline comparison score. Moreover, Claude-3.5-Sonnet outperforms Claude-3.7-Sonnet, the most expensive model, on the unweighted score and Top-1 accuracy. Although technical reports often emphasize gains from increased scale, our findings suggest that this relationship weakens for clinical pre‐consultation.

\noindent\textbf{Task-specific Fine-tuning matters.}\\
If model size does not guarantee pre‐consultation ability, what does? We hypothesize that once a model’s medical knowledge surpasses a certain threshold, its performance depends primarily on how effectively it can leverage that knowledge to generate appropriate questions. This interpretation is supported by the underperformance of Medgemma-4B, despite its presumed advantage in medical knowledge. To test this, we construct a 3k pre‐consultation dialogue dataset independent from EPAG—generated by LLMs and rigorously reviewed by clinical experts—and fine-tune Qwen-2.5 models (7B, 32B, 72B) using LoRA \citep{hu2021loralowrankadaptationlarge}. Figure \ref{fig:sft} compares each model’s performance before and after supervised fine-tuning. Consistent with our earlier analysis, the baseline models do not exhibit strict monotonic gains with size: while Top-1 accuracy improves as model size increases, the other three metrics rank as 32B < 7B < 72B. After SFT, all models show marked improvements across most metrics, with 32B benefiting the most. Although the base models fall below both the human expert and GPT-4.1, fine-tuned models often exceed the human expert—and notably, 7B and 32B match or even surpass GPT-4.1. Qwen2.5-72B's slight decline in Top-\textit{k} accuracy after fine-tuning possibly suggests underfitting, likely because our 3k-dialogue dataset is insufficient to fully optimize the largest model but more than adequate for the smallest model, making 32B the optimal size for this dataset. Overall, the peaking performance of fine-tuned Qwen2.5-32B demonstrates that relatively small open-source models, when trained on high-quality, task-specific data, can outperform larger, more expensive models in specialized applications. 

\noindent\textbf{Not all HPI directly lead to correct diagnosis.}\\
As shown in Figure \ref{fig:epag}, the amount of HPI increases with the number of dialogue turns, while diagnostic accuracy does not. Appendix \ref{appendix:analysis} exemplifies why more HPI does not directly correlate with accurate differential diagnosis. If a model fixates on certain keywords that are loosely connected to the correct diagnosis, it may ask numerous guideline-related but clinically less significant questions and even increase the likelihood of misdiagnosis.

\begin{figure}[t]
    \centering
    \includegraphics[width=0.95\columnwidth]{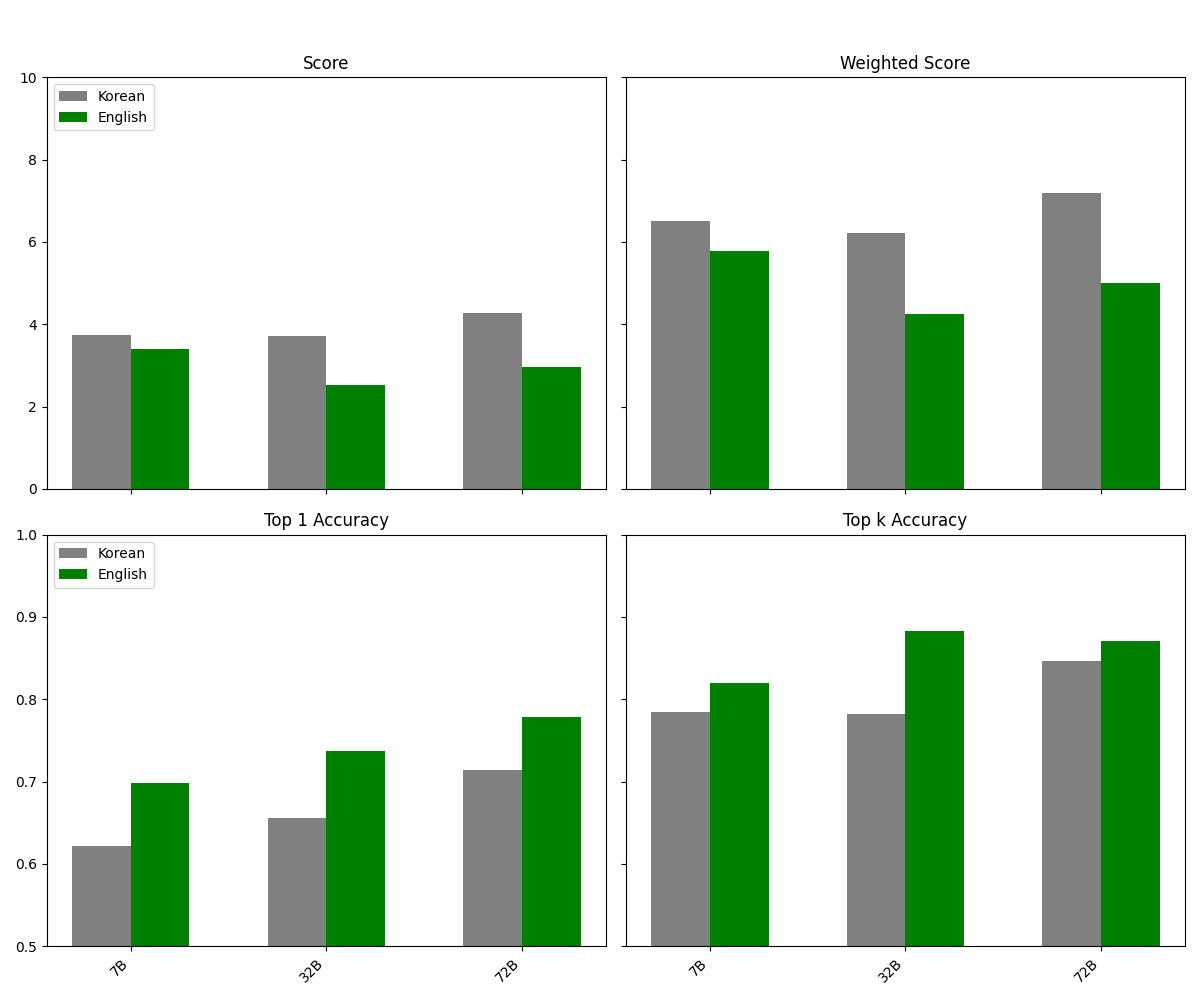}
    \caption{Performance of Qwen-2.5 models (7B, 32B, 72B) on Korean (grey) versus English (green) dialogues.}
    \label{fig:english}
\end{figure}

\noindent\textbf{Language affects dialogue patterns.}\\
With the prior experiments done in Korean, we explore whether the used language makes any difference by comparing English and Korean dialogues with Qwen 2.5 models (7B, 32B, 72B). We hypothesize that English pre-consultations would yield stronger performance as the English training corpus is understood to be much larger than Korean. Surprisingly, Figure \ref{fig:english} shows that Korean dialogues produce higher HPI-diagnostic guideline comparison scores, while English dialogues achieve superior disease diagnosis accuracy. A qualitative review explains this enigma: in English, the model frequently pursues deep, repetitive follow-ups on a single symptom—enhancing diagnostic confidence but generating fewer unique atomic units. By contrast, in Korean sessions it casts a wider net, querying a broader array of symptoms, which boosts HPI scores but dilutes focus and can introduce multiple diagnostic possibilities. This behavior aligns with our earlier finding that \emph{not all HPI directly lead to correct diagnosis.} 

\section{Conclusion}

We present \textbf{EPAG}, a benchmark dataset and automated pipeline for \textbf{E}valuating the \textbf{P}re-consultation \textbf{A}bility of LLMs using diagnostic \textbf{G}uidelines. Experiments show that model size does not guarantee performance, and not all extracted HPI contribute directly to diagnosis, highlighting the need for future research to quantify the impact of each HPI component on specific diagnosis and refine pre-consultation models. Additional studies demonstrate that smaller open-source LLMs can surpass larger proprietary models when fine-tuned with high-quality data, and that the language used during pre-consultation shapes dialogue characteristics.

\section*{Limitation and Future Work}

The EPAG benchmark dataset includes 26 diseases across 10 ICD-11 chapters but focuses solely on text-based pre-consultation models, excluding diseases that require physical test results, such as X-rays, or MRIs, which are more common in real-world settings. Therefore, future work should incorporate multi-modal evaluation of pre-consultation models to process inputs beyond text, including medical images.

\section*{Ethics Statement}

While our proposed evaluation pipeline for assessing the pre-consultation abilities of LLMs demonstrates a high correlation with human evaluation, it has limitations and does not cover all disease categories. As such, the experimental results presented in this paper should not be considered definitive. The selection of a model for any specific clinical application should involve thorough assessment before being deployed in practice.

\bibliography{anthology,custom}
\bibliographystyle{acl_natbib}

\appendix
\begin{appendix}

\newpage

\section{Source of Diagnostic Guidelines}
\label{appendix:source}

\begin{itemize}
    \item Cardiology: American College of Cardiology \footnote{https://www.acc.org/}, American Heart Association \footnote{https://www.heart.org/}
    \item Oncology: National Comprehensive Cancer Network \footnote{https://www.nccn.org/}
    \item Stroke: American Heart Association, American Stroke Association \footnote{https://www.stroke.org/en/}
    \item Allergy \& Immunology: Joint Task Force for Practice Parameters \footnote{https://www.aaaai.org/allergist-resources/statements-practice-parameters/practice-parameters-guidelines}, American Academy of Allergy Asthma \& Immunology \footnote{https://www.aaaai.org/}, American College of Allergy Asthma and Immunology \footnote{https://acaai.org/}
    \item Gastroenterology: American Gastroenterological Association Homepage \footnote{https://gastro.org/}
    \item HIV/AIDS: U.S. Preventive Services Task Force \footnote{https://www.uspreventiveservicestaskforce.org/uspstf/}
    \item Pulmonology: Global Initiative for Chronic Obstructive Lung Disease \footnote{https://goldcopd.org/}, Global Initiative for Asthma \footnote{https://ginasthma.org/}
    \item Nephrology: Improving Global Outcomes \footnote{https://kdigo.org/}
    \item Diabetes: American Diabetes Association \footnote{https://diabetes.org/}
    \item General Surgery: American College of Surgeons \footnote{https://www.facs.org/}
    \item Rheumatology: European League Against Rheumatism \footnote{https://www.eular.org/}
    \item Endocrinology: American Association of Clinical Endocrinologists \footnote{https://www.aace.com/}
\end{itemize}

\newpage

\section{Diagnostic Guideline Example}
\label{appendix:guideline}

\begin{table}[h]
\centering
\resizebox{\columnwidth}{!}{%
\begin{tabular}{|c|c|}
\hline
\textbf{Weight} & \textbf{Feature}                                              \\ \hline
high            & Palpable Breast Lump                                          \\ \hline
high            & Nipple Discharge, Bloody or Spontaneous                       \\ \hline
high            & Skin Changes: Peau d'orange, Ulceration, Erythema, Thickening \\ \hline
high            & New-Onset Nipple Inversion/Retraction                         \\ \hline
high            & Axillary Masses/Lymphadenopathy                               \\ \hline
medium          & Asymmetry in Breast Size/Shape, New Onset                     \\ \hline
medium          & Nipple/Areolar Eczema or Itching                              \\ \hline
medium          & Localized Thickening or Induration                            \\ \hline
medium          & Systemic Symptoms: Weight Loss, Fatigue, Night Sweats, Fever  \\ \hline
medium          & Pregnancy/Lactation-Related Abnormalities                     \\ \hline
medium          & Post-Surgical or Post-Radiation Breast Changes                \\ \hline
high            & Family History of Breast Cancer, BRCA Mutation                \\ \hline
high            & Genetic Predisposition: BRCA1/BRCA2, TP53, PALB2 etc.         \\ \hline
high            & Prior Biopsy with Atypia or LCIS/ADH                          \\ \hline
medium          & Hormonal Factors: Early Menarche, Late Menopause, HRT Use     \\ \hline
high            & Prior Chest Radiation Therapy, esp. 10$\sim$30 y/o            \\ \hline
\end{tabular}
}
\caption{Diagnostic guidelines for breast cancer.}
\end{table}

\section{Disease Categorization}
\label{appendix:disease}

To enhance the generalization and reliability of our benchmarking system, we adopt the International Classification of Diseases, 11th Revision (ICD-11) as the main categorization of diseases. This approach ensures comprehensive coverage across diverse disease groups. For better alignment with real-world clinical decision-making we assign each disease to a Primary Specialty and, where applicable, one or more Secondary Specialties.

\subsection{Primary Specialty Selection Criteria}

Each disease is assigned to a Primary Specialty, the leading specialty responsible for the disease's management, based on the following:

\begin{enumerate}
    \item ICD-11 Disease Classification:
    \begin{itemize}
        \item Each disease is mapped to its corresponding ICD-11 chapter, which indicates the major body system or disease category it belongs to.
        \item The specialty most commonly responsible for managing diseases in each chapter is assigned as the Primary Specialty.
    \end{itemize}
    \item International Clinical Guidelines: The Primary Specialty is further validated using well established medical guidelines from globally recognized organizations listed in Appendix \ref{appendix:source}. 
    \item Standard Medical Practice: The most commonly designated department responsible for managing the disease in hospitals and healthcare settings is selected.
\end{enumerate}

\subsection{Secondary Specialty Selection Criteria}

Many diseases require collaboration across multiple specialties. A Secondary Specialty, additional specialties that frequently contribute to diagnosis, treatment, or complication management, is assigned in cases where:

\begin{enumerate}
    \item Multidisciplinary care is essential.
    \begin{itemize}
        \item Conditions which require involvement from multiple specialties for optimal management.
        \item Example: Stroke (8B20)
        \begin{itemize}
            \item Primary: Neurology (acute treatment and long-term management)
            \item Secondary: Cardiology (stroke prevention in atrial fibrillation), Rehabilitation Medicine (post-stroke recovery)
        \end{itemize}
    \end{itemize}
    \item Complication management is required.
    \begin{itemize}
        \item Specialties involved in managing complications related to the primary disease.
        \item Example: Diabetes (5A14)
        \begin{itemize}
            \item Primary: Endocrinology (blood sugar control, metabolic regulation)
            \item Secondary: Nephrology (diabetic nephropathy), Cardiology (cardiovascular risk)
        \end{itemize}
    \end{itemize}
    \item Surgical vs. Non-Surgical considerations.
    \begin{itemize}
        \item Conditions where both medical and surgical specialties play a role.
        \item Example: Colorectal Cancer (2B91)
        \begin{itemize}
            \item Primary: Oncology (chemotherapy and cancer management)
            \item Secondary: Gastroenterology (diagnosis via colonoscopy), General Surgery (surgical treatment)
        \end{itemize}
    \end{itemize}

By structuring disease classification based on these criteria, we ensure that our benchmark system accurately represents real-world clinical workflows and enhances the applicability of AI-driven medical decision support tools.

\end{enumerate}

\section{Organized Unit Example}
\label{appendix:unit}

\textbf{Main Symptom:} \\
I keep coughing and have difficulty breathing. 

\subsection{[Question, Options, Answer] Triplet}

\textbf{Question:} When you cough, do you produce any sputum? \\
\textbf{Options:} Dry cough with no sputum, White or clear sputum, Yellow or green sputum, Red or brown sputum\\
\textbf{Answer:} White or clear sputum\\\\
\textbf{Question:} When is your difficulty breathing worse? \\
\textbf{Options:} I have difficulty breathing even when I am at rest, I have difficulty breathing when walking on flat ground, I have difficulty breathing when climbing stairs or going uphill, I only have difficulty breathing when I move quickly or exercise\\
\textbf{Answer:} I have difficulty breathing when walking on flat ground\\\\
\textbf{Question:} How long have you had the coughing and difficulty breathing symptoms? \\
\textbf{Options:} Less than 2 weeks, 2 weeks to 3 months, 3 months to 6 months, More than 6 months\\
\textbf{Answer:} More than 6 months\\\\
\textbf{Question:} Do you smoke? \\
\textbf{Options:} I currently smoke, I used to smoke but quit, I have never smoked, I am often exposed to secondhand smoke\\
\textbf{Answer:} I currently smoke\\\\
\textbf{Question:} Do you have any symptoms while sleeping? \\
\textbf{Options:} I wake up because I can't breathe, I can't sleep due to severe coughing, I need more than one pillow to breathe properly, I snore a lot, I sleep without any special symptoms\\
\textbf{Answer:} I wake up because I can't breathe

\subsection{Organized Units}
\begin{itemize}
    \item White or clear sputum is produced when coughing.
    \item The difficulty in breathing worsens when walking on flat ground.
    \item The coughing and difficulty in breathing symptoms have lasted more than 6 months.
    \item I currently smoke.
    \item I wake up during sleep because I can't breathe.
\end{itemize}

\section{Analysis}
\label{appendix:analysis}

The following case involves a patient expected to be diagnosed with \textit{Acute Kidney Injury}. MedGemma-4B is used as the doctor agent model.

\begin{table}[h]
\resizebox{\columnwidth}{!}{%
\begin{tabular}{l}

\toprule[1.3pt]
\textbf{Chief Complaint:} Decreased urine output and flank pain.                     \\\midrule[1.3pt]
\textbf{HPI from 5-turn dialogue}  \\\midrule

 There is pain in the right flank.\\
 The amount of urine has decreased.\\
 Recently had symptoms of a cold.\\
 Takes antihypertensive medication regularly.\\
 No history of urinary stones.\\\midrule

\textbf{Diagnosis:} \textit{Acute Kidney Injury} (correct)\\\midrule[1.2pt]

\textbf{HPI from 6-turn dialogue}\\\midrule

 There is pain in the right flank.\\
 The amount of urine has decreased.\\
 Recently had symptoms of a cold.\\
 Takes antihypertensive medication regularly.\\
 No history of urinary stones.\\
The flank pain is severe, rated 7 out of 10 in intensity. \color{Green}{(Added)}\\\midrule

\textbf{Diagnosis:} \textit{Renal Colic due to Urinary Stone} (incorrect)\\

\bottomrule[1.3pt]
\end{tabular}
}
\end{table}

\noindent Although both \textit{Acute Kidney Injury} and \textit{Renal Colic} can present with flank pain, the additional 6th turn provides patient information about the intensity of pain, which may have shifted the model’s diagnostic focus away from other relevant symptomatic information. \textit{Renal Colic} typically results from urinary stone, leading to severe pain. In this case, highlighting the severity of flank pain may have caused the model to prioritize pain-centric reasoning, which misled the differential diagnosis toward \textit{Renal Colic}. While the additional information (pain intensity) is clinically relevant and could aid a physician’s understanding, it may have inadvertently diverted the model's diagnostic focus.




\newpage

\begin{table*}[ht]
\centering
\resizebox{\textwidth}{!}{%
\begin{tabular}{c|c|c|c|c}

    \toprule[1.3pt]
    \textbf{ICD-11 Chapter}                                                                                     & \textbf{Disease}                                                                              & \textbf{ICD-11 Code} & \textbf{Primary Specialty} & \textbf{Secondary Specialty}                                                                                          \\ \midrule[1.3pt]
    \multirow{8}{*}{Neoplasms}                                                                                  & Breast Cancer                                                                                 & 2E65                 & Oncology                   & General Surgery                                                                                                       \\ \cline{2-5} 
                                                                                                                & Prostate Cancer                                                                               & 2C82                 & Oncology                   & Urology                                                                                                               \\ \cline{2-5} 
                                                                                                                & Colorectal Cancer                                                                             & 2B91                 & Oncology                   & \begin{tabular}[c]{@{}c@{}}Gastroenterology, \\ General Surgery\end{tabular}                                          \\ \cline{2-5} 
                                                                                                                & Lung Cancer                                                                                   & 2C25                 & Oncology                   & \begin{tabular}[c]{@{}c@{}}Pulmonology, \\ Thoracic Surgery\end{tabular}                                              \\ \cline{2-5} 
                                                                                                                & Gastric Cancer                                                                                & 2B72                 & Oncology                   & \begin{tabular}[c]{@{}c@{}}Gastroenterology, \\ General Surgery\end{tabular}                                          \\ \midrule[1.3pt]
    \multirow{7}{*}{\begin{tabular}[c]{@{}c@{}}Diseases of the \\ Circulatory System\end{tabular}}              & Hypertrophic Cardiomyopathy                                                                   & BC43.1               & Cardiology                 & Medical Genetics                                                                                                      \\ \cline{2-5} 
                                                                                                                & Peripheral Artery Disease                                                                     & BD4Z                 & Cardiology                 & Vascular Surgery                                                                                                      \\ \cline{2-5} 
                                                                                                                & Atrial Fibrillation                                                                           & BC81.3               & Cardiology                 & \begin{tabular}[c]{@{}c@{}}Neurology \\ (Stroke Risk), \\ Internal Medicine\end{tabular}                              \\ \cline{2-5} 
                                                                                                                & Heart Failure                                                                                 & BD1Z                 & Cardiology                 & \begin{tabular}[c]{@{}c@{}}Endocrinology \\ (Diabetes-related)\end{tabular}                                           \\ \midrule[1.3pt]
    \multirow{3}{*}{\begin{tabular}[c]{@{}c@{}}Diseases of the \\ Nervous System\end{tabular}}                  & Stroke                                                                                        & 8B20                 & Neurology                  & \begin{tabular}[c]{@{}c@{}}Cardiology, \\ Rehabilitation Medicine\end{tabular}                                        \\ \cline{2-5} 
                                                                                                                & \begin{tabular}[c]{@{}c@{}}Aneurysmal Subarachnoid \\ Haemorrhage\end{tabular}                & 8B01.0               & Neurology                  & \begin{tabular}[c]{@{}c@{}}Neurosurgery, \\ Emergency Medicine\end{tabular}                                           \\ \midrule[1.3pt]
    \multirow{9}{*}{\begin{tabular}[c]{@{}c@{}}Diseases of the \\ Immune System\end{tabular}}                   & Anaphylaxis                                                                                   & 4A84                 & Allergy \& Immunology      & Emergency Medicine                                                                                                    \\ \cline{2-5} 
                                                                                                                & Systemic Sclerosis                                                                            & 4A42                 & Rheumatology               & \begin{tabular}[c]{@{}c@{}}Pulmonology \\ (Lung fibrosis), \\ Cardiology \\ (Cardiac involvement)\end{tabular}        \\ \cline{2-5} 
                                                                                                                & \begin{tabular}[c]{@{}c@{}}Systemic Lupus \\ Erythematosus\end{tabular}                       & 4A40.0               & Rheumatology               & \begin{tabular}[c]{@{}c@{}}Nephrology \\ (Lupus Nephritis), \\ Cardiology\\ (Vascular Complications)\end{tabular}     \\ \midrule[1.3pt]
    \begin{tabular}[c]{@{}c@{}}Diseases of the \\ Skin\end{tabular}                                             & Atopic Dermatitis                                                                             & EA80                 & Allergy \& Immunology      & Dermatology                                                                                                           \\ \midrule[1.3pt]
    \multirow{6}{*}{\begin{tabular}[c]{@{}c@{}}Diseases of the \\ Digestive System\end{tabular}}                & Ulcerative Colitis                                                                            & DD71                 & Gastroenterology           & \begin{tabular}[c]{@{}c@{}}Rheumatology \\ (Autoimmune-related)\end{tabular}                                          \\ \cline{2-5} 
                                                                                                                & \begin{tabular}[c]{@{}c@{}}Nonalcoholic Fatty \\ Liver Disease\end{tabular}                   & DB92.Z               & Gastroenterology           & \begin{tabular}[c]{@{}c@{}}Endocrinology \\ (Metabolic Syndrome)\end{tabular}                                         \\ \cline{2-5} 
                                                                                                                & \begin{tabular}[c]{@{}c@{}}Irritable Bowel Syndrome \\ with Constipation (IBS-C)\end{tabular} & DD91.00              & Gastroenterology           & \begin{tabular}[c]{@{}c@{}}Psychiatry \\ (Stress-related IBS)\end{tabular}                                            \\ \cline{2-5} 
                                                                                                                & Acute Pancreatitis                                                                            & DC31                 & Gastroenterology           & General Surgery                                                                                                       \\ \midrule[1.3pt]
    \begin{tabular}[c]{@{}c@{}}Certain Infectious \\ or \\ Parasitic Diseases\end{tabular}                      & \begin{tabular}[c]{@{}c@{}}Human Immunodeficiency \\ Virus (HIV) Infection\end{tabular}       & 1C62                 & Infectious Diseases        & Immunology                                                                                                            \\ \midrule[1.3pt]
    \multirow{4}{*}{\begin{tabular}[c]{@{}c@{}}Diseases of the \\ Respiratory System\end{tabular}}              & \begin{tabular}[c]{@{}c@{}}Chronic Obstructive \\ Pulmonary Disease\end{tabular}              & CA22                 & Pulmonology                & Internal Medicine                                                                                                     \\ \cline{2-5} 
                                                                                                                & Asthma                                                                                        & CA23                 & Pulmonology                & Allergy \& Immunology                                                                                                 \\ \cline{2-5} 
                                                                                                                & Allergic Rhinitis                                                                             & CA08.0               & Allergy \& Immunology      & \begin{tabular}[c]{@{}c@{}}Otorhinolaryngology, \\Pulmonology\end{tabular}                                            \\ \midrule[1.3pt]
    \begin{tabular}[c]{@{}c@{}}Diseases of the \\ Genitourinary System\end{tabular}                             & Acute Kidney Injury                                                                           & GB60                 & Nephrology                 & Critical Care Medicine                                                                                                \\ \midrule[1.3pt]
    \multirow{5}{*}{\begin{tabular}[c]{@{}c@{}}Endocrine, Nutritional \\ or \\ Metabolic Diseases\end{tabular}} & Diabetes Mellitus                                                                             & 5A14                 & Endocrinology              & \begin{tabular}[c]{@{}c@{}}Nephrology \\ (Diabetes-related \\ Kidney Disease)\end{tabular}                            \\ \cline{2-5} 
                                                                                                                & Hypothyroidism                                                                                & 5A00                 & Endocrinology              & \begin{tabular}[c]{@{}c@{}}Cardiology \\ (Atrial Fibrillation Risk), \\ Psychiatry \\ (Depression Link)\end{tabular}  \\ \bottomrule[1.3pt]
    \end{tabular}%
}
\caption{List of 26 diseases consisting EPAG benchmark. Detailed classification of diseases including ICD-11 Chapter, ICD-11 Code, Primary Specialty, and Secondary Specialty are provided.}
\label{tab:disease_all}
\end{table*}

\begin{table*}[t]
\centering
\resizebox{\textwidth}{!}{%
\begin{tabular}{|ccl|}
\hline
\multicolumn{3}{|c|}{\textbf{Patient Profile}}                                                                                                                                                                                        \\ \hline
\multicolumn{1}{|c|}{\textbf{Disease Name}}                                & \multicolumn{2}{c|}{\textbf{Breast Cancer}}                                                                                                              \\ \hline
\multicolumn{1}{|c|}{\textbf{Typicality}}                                  & \multicolumn{2}{c|}{\textbf{Normal}}                                                                                                                     \\ \hline
\multicolumn{1}{|c|}{\multirow{4}{*}{\textbf{Basic Information}}}          & \multicolumn{1}{c|}{\textbf{Age}}                           & 51                                                                                         \\ \cline{2-3} 
\multicolumn{1}{|c|}{}                                                     & \multicolumn{1}{c|}{\textbf{Sex}}                           & Female                                                                                     \\ \cline{2-3} 
\multicolumn{1}{|c|}{}                                                     & \multicolumn{1}{c|}{\textbf{Height}}                        & 162cm                                                                                      \\ \cline{2-3} 
\multicolumn{1}{|c|}{}                                                     & \multicolumn{1}{c|}{\textbf{Weight}}                        & 62kg                                                                                       \\ \hline
\multicolumn{1}{|c|}{\multirow{8}{*}{\textbf{History of Present Illness}}} & \multicolumn{1}{c|}{\textbf{Location}}                      & Left breast and adjacent axillary region                                                   \\ \cline{2-3} 
\multicolumn{1}{|c|}{}                                                     & \multicolumn{1}{c|}{\textbf{Quality}}                       & Firm, irregular mass                                                                       \\ \cline{2-3} 
\multicolumn{1}{|c|}{}                                                     & \multicolumn{1}{c|}{\textbf{Severity}}                      & 4/10 (Mild pain but significant anxiety)                                                   \\ \cline{2-3} 
\multicolumn{1}{|c|}{}                                                     & \multicolumn{1}{c|}{\textbf{Duration}}                      & Approximately 3 months                                                                     \\ \cline{2-3} 
\multicolumn{1}{|c|}{}                                                     & \multicolumn{1}{c|}{\textbf{Timing}}                        & Slight variations with menstrual cycle, discovered accidentally during routine examination \\ \cline{2-3} 
\multicolumn{1}{|c|}{}                                                     & \multicolumn{1}{c|}{\textbf{Context}}                       & Detected by the patient herself during a routine breast examination                        \\ \cline{2-3} 
\multicolumn{1}{|c|}{}                                                     & \multicolumn{1}{c|}{\textbf{Modifying Factors}}             & Slight reduction in swelling post-menstruation, no specific alleviating factors            \\ \cline{2-3} 
\multicolumn{1}{|c|}{}                                                     & \multicolumn{1}{c|}{\textbf{Associated Signs and Symptoms}} & Mild nipple discharge, slight fatigue, minimal pain                                        \\ \hline
\multicolumn{1}{|c|}{\multirow{4}{*}{\textbf{Additional Information}}}     & \multicolumn{1}{c|}{\textbf{Family History}}                & No family history of breast cancer or similar cancers                                      \\ \cline{2-3} 
\multicolumn{1}{|c|}{}                                                     & \multicolumn{1}{c|}{\textbf{Previous Surgery or Illness}}   & No previous history of breast-related surgery or conditions                                \\ \cline{2-3} 
\multicolumn{1}{|c|}{}                                                     & \multicolumn{1}{c|}{\textbf{Lifestyle Changes}}             & No recent changes in lifestyle; the patient aims for early detection through screening     \\ \cline{2-3} 
\multicolumn{1}{|c|}{}                                                     & \multicolumn{1}{c|}{\textbf{Health Check-ups}}              & Regularly undergoes women's health check-ups                                               \\ \hline
\multicolumn{1}{|c|}{\multirow{2}{*}{\textbf{Pain Area}}}                  & \multicolumn{2}{l|}{Left chest (pectoral region)}                                                                                                        \\ \cline{2-3} 
\multicolumn{1}{|c|}{}                                                     & \multicolumn{2}{l|}{Left anterior acromio-clavicular region}                                                                                             \\ \hline
\multicolumn{1}{|c|}{\multirow{2}{*}{\textbf{Past Medical History}}}       & \multicolumn{2}{l|}{No history of breast diseases}                                                                                                       \\ \cline{2-3} 
\multicolumn{1}{|c|}{}                                                     & \multicolumn{2}{l|}{No other chronic illnesses}                                                                                                          \\ \hline
\multicolumn{1}{|c|}{\multirow{3}{*}{\textbf{Social History}}}             & \multicolumn{2}{l|}{Office worker, full-time}                                                                                                             \\ \cline{2-3} 
\multicolumn{1}{|c|}{}                                                     & \multicolumn{2}{l|}{Non-smoker, drinks alcohol 1-2 times per week}                                                                                       \\ \cline{2-3} 
\multicolumn{1}{|c|}{}                                                     & \multicolumn{2}{l|}{Regular health check-ups and breast self-examination}                                                                                \\ \hline
\multicolumn{1}{|c|}{\textbf{Chief Complaint}}                               & \multicolumn{2}{l|}{A firm lump in the left chest, causing anxiety}                                                                                      \\ \hline
\end{tabular}
}%
\caption{Sample patient profile with breast cancer.}
\label{tab:sample_patient}
\end{table*}

\begin{figure*}[h]
    \centering
    \includegraphics[width=\textwidth]{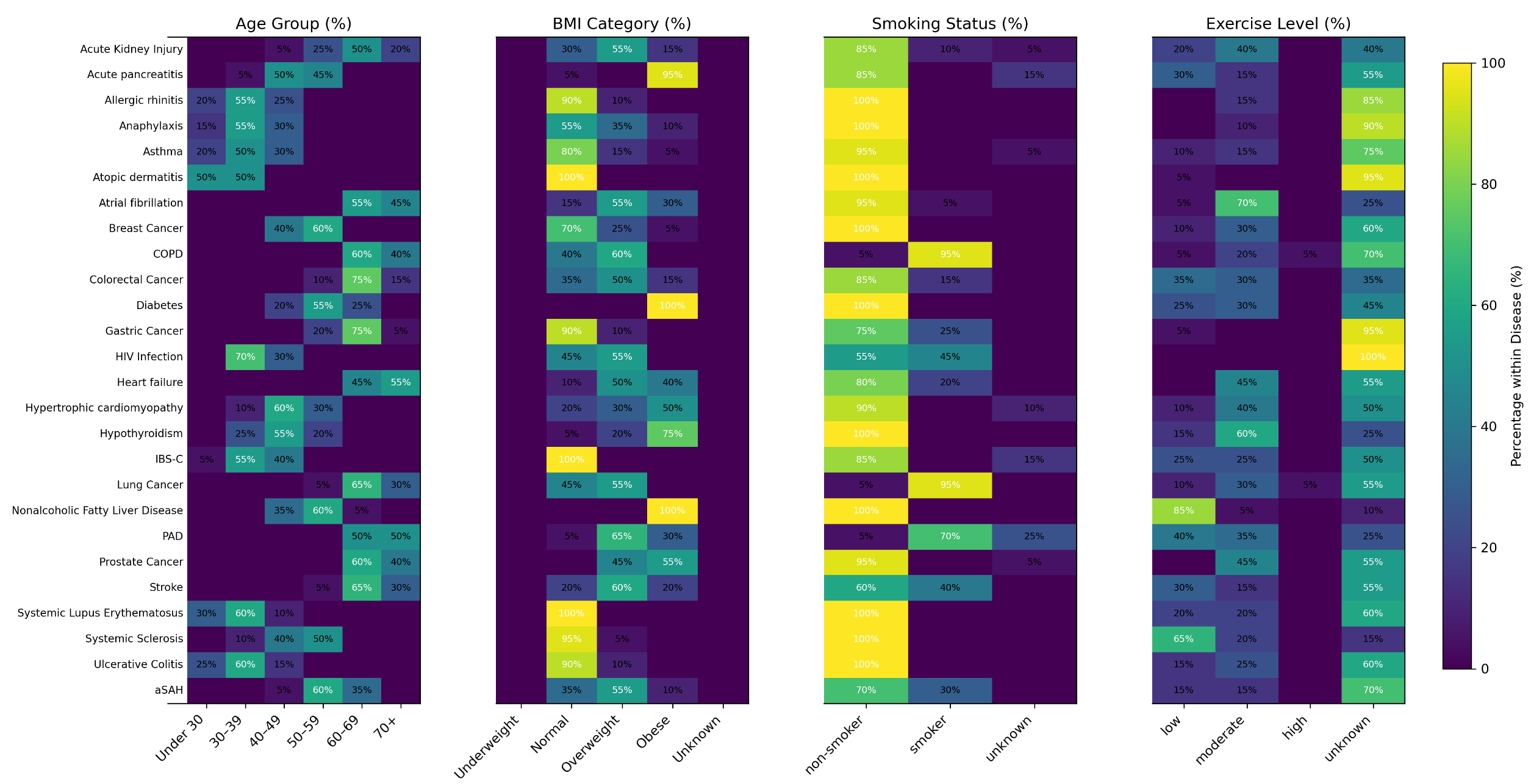}
    \caption{Distribution of age group, BMI category, smoking status, exercise level of patients for each disease.}
    \label{fig:diversity}
\end{figure*}

\begin{table*}[t]
\centering
\resizebox{1.0\textwidth}{!}{%
\begin{tabular}{l}
\toprule[1.3pt]
\textbf{Original}                                                                     \\ \toprule[1.3pt]
\begin{tabular}[c]{@{}l@{}}당신은 아래와 같은 프로필을 가진 환자입니다. \\ \textcolor{gray}{\{patient\_information\}}\\ \\ 적절한 진단을 위해 의사의 문진에 충실히 대답해야 합니다. 의사가 제시한 질문지 중 한 가지를 골라 대답하세요.\\ \\ 출력 형식:\\ Answer:\end{tabular}                                                                                                          \\ \toprule[1.3pt]
\textbf{Translated}                                                                 \\ \toprule[1.3pt]
\begin{tabular}[c]{@{}l@{}}You are a patient with the following profile:\\ \textcolor{gray}{\{patient\_information\}}\\ \\ You should faithfully answer the doctor's inquiries for an appropriate diagnosis. Choose one of the questions provided by the doctor and respond.\\ \\ Output format:\\ Answer:\end{tabular} \\ \toprule[1.3pt]
\end{tabular}
}
\caption{Patient Agent Prompt}
\label{tab:patient_prompt}
\end{table*}

\begin{table*}[]
\centering
\resizebox{\textwidth}{!}{%
\begin{tabular}{l}
\toprule[1.3pt]
\textbf{Original}   \\ \toprule[1.3pt]

**Prompt:**\\
Below is a set of consultation dialogues between a doctor and a patient with {disease}. The dialogue consists of the patient's chief complaint, a few turns of questions, options, and answer triplets. \\The questions and options are given by the doctor, and the answers are provided by the patient. \\

Your task is to organize the dialogue in a clear, information-based manner using bullet points. Each bullet point should contain only one piece of information. \\This structured information is essential for diagnosing the patient's condition, so make sure to extract as much relevant information as possible.\\\\

**Guidelines for Organizing:**\\
- Do not include the main symptom in the bullet points. \\The main symptom is just for reference and should not be summarized in bullet points.\\
- Focus only on the information that can be inferred from the Question-Options-Answer triplets.\\
- Each bullet point must present only one piece of information.\\
- Avoid sentences with multiple clauses. \\For example, instead of “The patient has cough and sputum,” break it down into two points:\\ “The patient has a cough” and “The patient has sputum.”\\
- Avoid using demonstrative pronouns (e.g., “this symptom”) and pronouns (e.g., “he/she”). Focus on the specific symptoms.\\
- Organize the information from the patient’s perspective, avoiding the doctor as the subject.\\
- Keep the language neutral and concise, stating only the facts shared by the patient.\\
- If the question asks about additional symptoms beyond the main symptom and the answer is that no other symptoms are present, \\list each symptom option provided in the question and state that the patient does not have each of those symptoms. \\For example, instead of "There are no other symptoms," specify each of the symptom option provided:\\ "There is no family history," "There is no weight loss," "There is no fever."\\
- Be precise and specific in organizing the information.\\ For example, if a question asks about "whether the patient has ever had any tests related to lumps or breasts," \\and the answer is "No," do not simply write "The patient has not had any tests." \\Instead, write, "The patient has not had any tests related to lumps or breasts."\\\\

**Example**:\\

\textcolor{gray}{\{examples\}}\\\\

**Input**\\

\textcolor{gray}{\{input\}}\\\\

**Organized Information:**\\
-

                    \\ \bottomrule[1.3pt]
\end{tabular}
}
\caption{Organizer Prompt}
\label{tab:organizer_prompt}
\end{table*}

\begin{table*}[]
\centering
\resizebox{\textwidth}{!}{%
\begin{tabular}{l}
\toprule[1.3pt]
\textbf{Original}   \\ \toprule[1.3pt]

You are a medical/health expert. Below is a conversation between a {disease} patient and a doctor.\\ In this case, evaluate whether [the interview conversation (A)] effectively leads to the [key diagnostic elements (B)], which are pre-defined for specific diseases. \\Here, (B) includes not only symptoms but also important elements such as past medical history, family history, and other disease diagnoses. \\

First, identify if (A) is relevant enough to {disease} and helpful in drawing out new information to diagnose {disease} given (H).\\ If not, output "Irrelevant/Redundant." \\If (A) is relevant to {disease} and helpful in drawing out new information to diagnose {disease} given (H), \\determine whether each item in (B) can be identified through the interview conversation (A). \\If two or more (B) items can be identified from (A), output the most relevant (B) item. If no (B) items can be identified through (A), output "None of above."\\\\

<Explanation of the provided information>\\
- **Dialogue History (H)**  \\
This is a prior conversation between the patient and the doctor. \\It includes the main symptom the patient reported, the questions the doctor asked to make a diagnosis, the options presented, and the patient's answer. \\Sometimes only the main symptom the patient complained about may be provided.\\
  
- **Interview Conversation (A)**  \\
This consists of the questions and options the doctor asks the patient for diagnostic purposes. \\The patient chooses one option from the given choices to respond.\\

- **Pre-defined Key Diagnostic Elements List (B)**  \\
Example: Persistent Cough, Hemoptysis (Coughing up Blood), Dyspnea (Shortness of Breath), Chest Pain, Unexplained Weight Loss, Family History of Lung Cancer, Smoking History, etc.\\\\

<Important Notes>  \\
1. **Evaluation Criteria**  \\
- Check if the interview conversation (A) is designed to identify (B), \\or if it directly helps to determine specific aspects of (B) such as the onset, duration, more exact location and frequency of symptoms.\\
- If (A) is related to an item in (B) but deviates from the patient disease which is {disease}, then output "None of above."\\
- Assess if the questions and options in (A) can effectively elicit relevant information related to (B) from the patient.\\
  
2. **Output Criteria**  
- Provide a brief Reason for\\ whether (A) can effectively elicit (B)-related information. Do not repeat the questions and options.\\
- The Reason should be up to two sentences.\\
- The Final Response should be either [(B) item] or "None of above." or "Irrelevant/Redundant."\\
- If multiple (B) items can be identified from (A), output only the one most directly related to (A).\\ If the relevance is judged to be the same, separate the related (B) items using "[OR]" and output them all.\\
- (H) is for reference only, so the evaluation should focus on whether (A) is related to (B).\\\\

**Example**:\\
\textcolor{gray}{\{example\}}\\\\

(H):\\
\textcolor{gray}{\{h\}}\\

(A): \\
\textcolor{gray}{\{a\}}\\

(B):\\
\textcolor{gray}{\{b\}}\\

Reason: \\

                    \\ \bottomrule[1.3pt]
\end{tabular}
}
\caption{Comparer Prompt}
\label{tab:comparer_prompt}
\end{table*}

\begin{table*}[t]
\centering
\resizebox{1.0\textwidth}{!}
{%
\begin{tabular}{l}
\toprule[1.3pt]
\textbf{Original}                                                                     \\ \toprule[1.3pt]
\begin{tabular}[c]{@{}l@{}}

You are a medical expert. Given 'patient\_info' and 'medical\_history', output the suspected disease names in order of highest probability. Output your prediction in English in YAML format.\\
\\
Instructions:\\
- Use only specific disease names related to the patient's symptoms.\\
- Prioritize based on main symptoms, severity, duration, and answers given in the medical history.\\
- Exclude diseases that don’t match the responses or are too generic.\\
- List the diseases in order of highest probability first.\\
- Do not provide any extra explanation.\\
\\
Output format:\\
Diseases:\\
- (probable diseases)\\

\end{tabular}                                                                                                          \\ \toprule[1.3pt]

\end{tabular}
}
\caption{Diagnostician Agent Prompt}
\label{tab:diagnostician_prompt}
\end{table*}

\begin{table*}[t]
\centering
\resizebox{1.0\textwidth}{!}
{%
\begin{tabular}{l}
\toprule[1.3pt]
\textbf{Original}                                                                     \\ \toprule[1.3pt]
\begin{tabular}[c]{@{}l@{}}

You are a medical expert. Given 'model\_predictions' and 'golden\_standard', decide if the predictions are correct. Output your reasoning in English in YAML format.\\\\
    
Instructions:\\
- Accept if the predicted disease is very similar to the actual one.\\
- Accept synonyms or other expressions for the same disease.\\
- Accept if the disease names include hierarchical (superior/inferior) relationships.\\
- Accept medical abbreviations as equivalent to official names.\\
- Allow regional/cultural expression differences.\\
- If at least one prediction is correct, consider it acceptable.\\
\\
Output format:\\
Reasoning: |\\
  (your reasoning in English)\\
Result: True/False\\

\end{tabular}                                                                                                          \\ \toprule[1.3pt]


\end{tabular}
}
\caption{Evaluator Prompt}
\label{tab:evaluator_prompt}
\end{table*}

\newpage

\begin{figure*}[h]
    \centering
    \includegraphics[width=\textwidth]{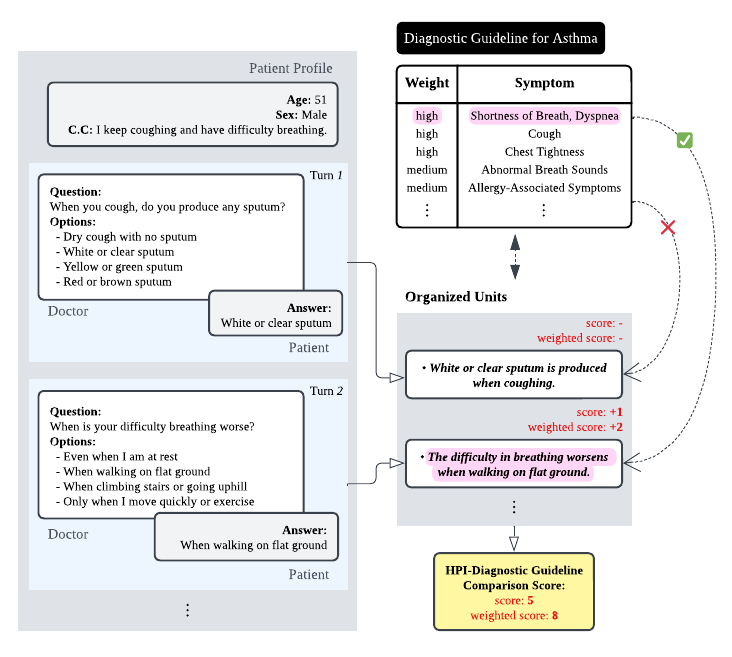}
    \caption{Sample pre-consultation dialogue and HPI-diagnostic guideline comparison process. Given basic patient information, including the chief complaint, the doctor asks questions and the patient selects answers from provided options. The dialogues are organized into atomic units, each of which is compared against a pre-defined diagnostic guideline. Units matching the guideline receive a score; those that do not are not scored.
}
    \label{fig:dialogue_appendix}
\end{figure*}

\begin{figure*}[t]
    \centering
    \includegraphics[width=0.9\textwidth]{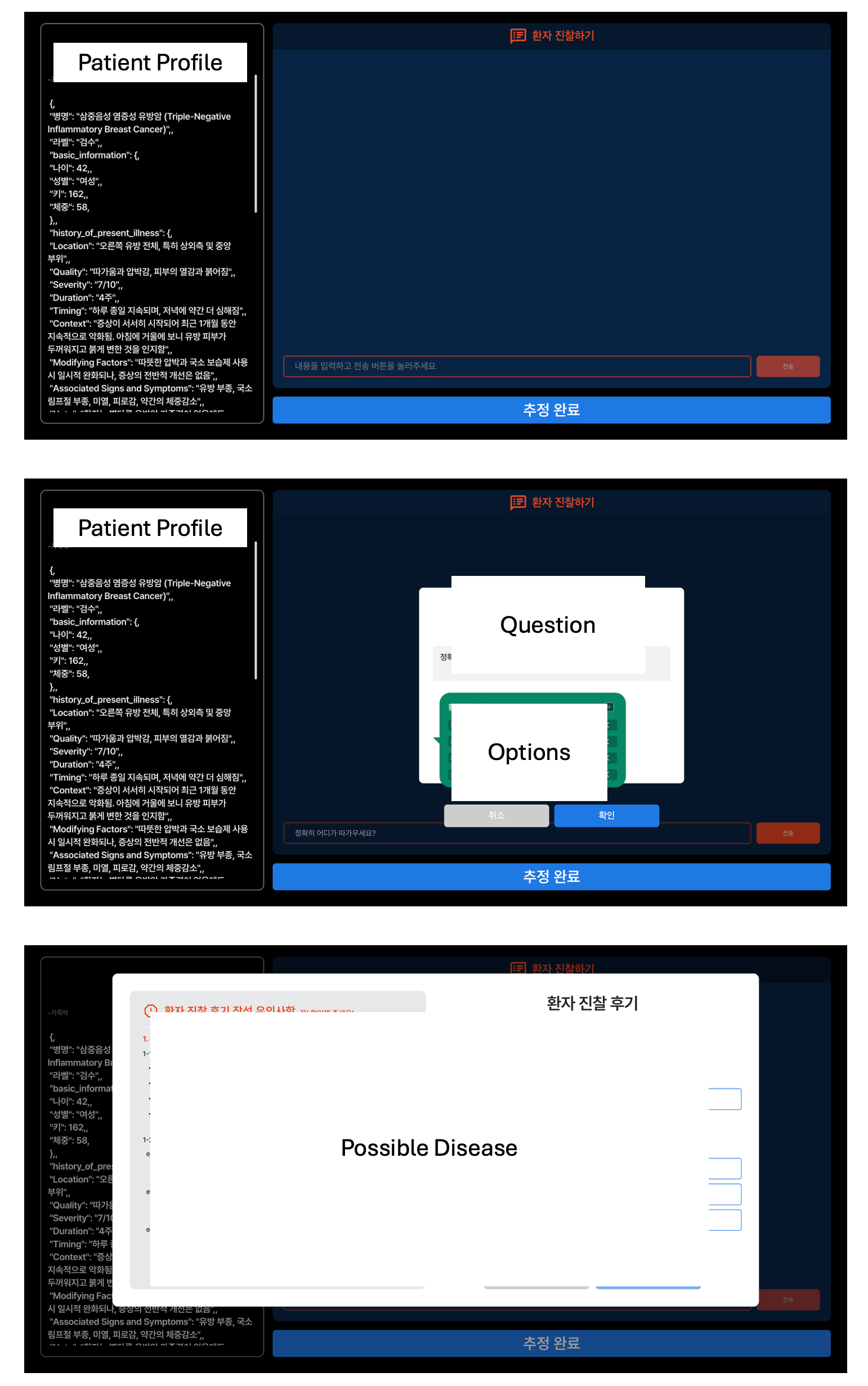}
    \caption{User interface used by human clinicians to simulate pre-consultation dialogues with patient agents. Given the patient profile displayed on the left, clinicians generate questions and response options for the patient agent to select. After each submission, the selected option is shown to the clinician, who then formulates the next question and options. After a series of dialogue turns, clinicians provide a diagnosis of the possible diseases.}
    \label{fig:human_screen}
\end{figure*}

\end{appendix}

\end{document}